%% file: main-arxiv.tex
\documentclass[3p,preprint]{elsarticle}

\usepackage{hyperref}
\journal{Journal of Web Semantics}

\usepackage{amsmath}
\usepackage{amssymb}
\usepackage{mathtools}
\usepackage{adjustbox}
\usepackage{booktabs}
\usepackage{algorithm}
\usepackage{algpseudocode}

\newcommand\ents{\ensuremath{\textsc{entities}}}
\newcommand\rels{\ensuremath{\textsc{relations}}}
\newcommand\dom{\ensuremath{\textsc{domain}}}
\newcommand\range{\ensuremath{\textsc{range}}}

\newcommand*\Let[2]{\State #1 $\gets$ #2}
\algrenewcommand\alglinenumber[1]{
    {\sf\tiny#1}}
\algrenewcommand\algorithmicrequire{\textbf{Precondition:}}

\begin{document}

\begin{frontmatter}

\title{Benchmarking neural embeddings for link prediction in knowledge graphs under semantic and structural changes}
% \tnotetext[mytitlenote]{Fully documented templates are available in the elsarticle package on \href{http://www.ctan.org/tex-archive/macros/latex/contrib/elsarticle}{CTAN}.}

% \address{Radarweg 29, Amsterdam}
% \fntext[myfootnote]{Since 1880.}

\author[mymainaddress]{Asan Agibetov\corref{mycorrespondingauthor}}
\cortext[mycorrespondingauthor]{Corresponding author}
\ead{asan.agibetov@meduniwien.ac.at}
\ead[url]{Code is available open source https://github.com/plumdeq/neuro-kglink}

\author[mymainaddress]{Matthias Samwald}

\address[mymainaddress]{Section for Artificial Intelligence and Decision Support, Medical University of Vienna, Austria}
% \address[mysecondaryaddress]{360 Park Avenue South, New York}

\begin{abstract} Recently, link prediction algorithms based on neural embeddings have gained tremendous popularity in the Semantic Web community, and
    are extensively used for knowledge graph completion. While algorithmic advances have strongly focused on efficient ways of learning embeddings,
    fewer attention has been drawn to the different ways their performance and robustness can be evaluated. In this work we propose an open-source
    evaluation pipeline, which benchmarks the accuracy of neural embeddings in situations where knowledge graphs may experience semantic and structural changes. We
    define relation-centric connectivity measures that allow us to connect the link prediction capacity to the structure of the knowledge graph.
    Such an evaluation pipeline is especially important to simulate the accuracy of embeddings for knowledge graphs that are expected to be frequently updated.

    %  
    % We focus on the link prediction problem for knowledge graphs, which is treated herein as a binary classification task on shallow knowledge graph
    % embeddings. By comparing, combining and extending different methodologies for link prediction on graph-based data coming from different domains,
    % we formalize a unified methodology for the quality evaluation benchmark of shallow neural embeddings for knowledge graphs. This benchmark is
    % then used to empirically investigate the potential of training neural generalized embeddings once for the entire graph and test them on all
    % relations, as opposed to the state-of-the-art way of training specialized embeddings one relation at a time. This new way of training the neural
    % embeddings and evaluating their quality is important for scalable link prediction with limited data. We perform extensive statistical repeated
    % sub-sample validation to empirically support our claims, and derive relation-centric connectivity measures for knowledge graphs to explain our
    % findings. Our evaluation pipeline is made open source, and with this we aim to draw more attention of the community towards an important issue
    % of transparency and reproducibility of neural embeddings evaluations. 
\end{abstract}


\end{frontmatter}

%\linenumbers

\input{./sections/intro.tex}
%
\input{./sections/methods.tex}
\input{./sections/structural_change.tex}
%
\input{./sections/experiments.tex}
\input{./sections/conclusion.tex}

\section*{Acknowledgements}

The computational results presented have been achieved in part using the Vienna Scientific Cluster (VSC).

%\section*{References}

\bibliography{biblio}

\end{document}

%% file: sections/intro.tex
\section{Introduction} \label{sec:intro} 

\begin{figure}[!h]
     \begin{center}
         \includegraphics[width=.4\linewidth]{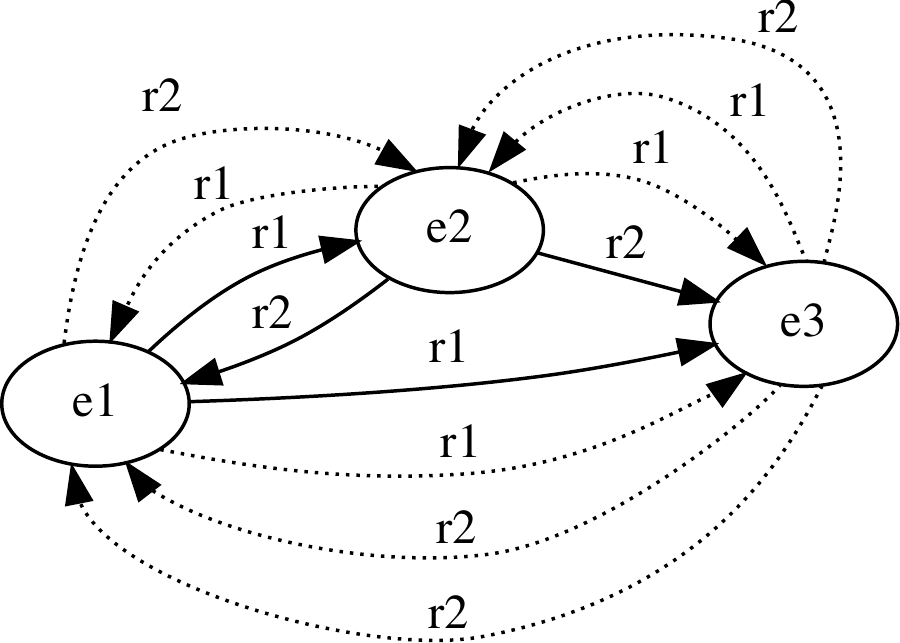}
     \end{center}
     \caption{\label{fig:kg-completion} A sample KG with three entities and two relation types. Positive links are drawn in bold, negative in dotted edge styles.}
\end{figure}

Link prediction, in general, is a problem of finding the missing or unknown links among inter-connected entities. This assumes that entities and links
can be represented as a graph, where entities are nodes and links (symmetric relationships) are edges (arcs if relationships are asymmetric). This
prediction problem has been most probably defined for the first time in the social network analysis community~\cite{libennowell_2003}, however, it has
soon become an important problem in other domains, and in particular in large-scale knowledge-bases~\cite{nickel_2016}, where it is used to add
missing data and discover new facts. When we are dealing with the link prediction problem for knowledge-bases, the semantic information contained
within is usually encoded as a knowledge graph (KG)~\cite{sri_2008}. For the purpose of this manuscript, we treat a knowledge graph as a graph where
links may have different types, and we conform to the \emph{closed-world} assumption. This means that all the existing (asserted) links are considered
\emph{positive}, and all the links which are unknown, and obtained via knowledge graph completion, are considered \emph{negative}
(Figure~\ref{fig:kg-completion}). 

% For what follows, consider a knowledge graph $KG$, consisting of 3 entities $e_1, e_2, e_3$ interconnected with links of 2 types $r_1$ and $r_2$, as
% depicted in . In the literature we stick to the close-world assumption, and consider all the possible knowledge graph completions as negative links.
% In Figure~\ref{fig:kg-completion} bold arcs indicate the known links (positive), and the dotted arcs the unknown (negative).

This separation into positive and negative links (examples) naturally allows us to treat the link prediction problem as a supervized classification problem with
binary predictors. However, while this separation enables a wealthy body of available and well-studied machine learning algorithms to be used for link prediction, the
main challenge is how to find the best representations for links. And this is the core subject of the recent research trend in learning suitable representations for
knowledge graphs, largely dominated by the so-called \emph{neural embeddings} (initially introduced for language modelling~\cite{mikolov_2013}). Neural embeddings are
numeric representations of nodes, and/or relations of the knowledge graph, in some continuous and dense vector space. These \emph{embeddings} are \emph{learned} with
neural networks by optimizing a specific objective function. Usually, the objective function models the constraints that neighboring nodes are \emph{embedded} close
to each other, and the nodes that are not directly connected, or separated via long paths in the graph, are embedded to stay far apart. A link in a knowledge graph is
then represented as a combination of node and/or relation type embeddings~\cite{nickel_2016,goyal_2018}. 

\subsection{Benchmarking accuracy and robustness of neural embeddings}

There are two major ways of measuring the accuracy of embeddings of entities in a knowledge graph for link prediction tasks, inspired by two different
fields: information retrieval~\cite{bordes_2013,distmult2015,complex2016,nickel_2016,kadlec_2017,wu_2017,nickel_2017} and graph-based data
mining~\cite{perozzi_2014,grover_2016,garciagasulla_2016,chamberlain_2017,alshahrani_2017,agibetov_semdeep_2018}. Information retrieval inspired
approaches seem to favor node-centric evaluations, which measure how well the embeddings are able to reconstruct the \emph{immediate} neighbourhood
for each node in the graph; these evaluations are based on the mean rank measurement and its variants (mean average precision, top $k$ results, mean
reciprocal rank). And graph-based data mining approaches tend to measure link-equality by recurring to the evaluation measurements such as, ROC
AUC and F-measure. See~\cite{crichton_2018} for more discussion on node- and link-equality measures. 

Besides evaluating the accuracy, some works have focused their attention on issues that might hinder the appropriate evaluation of embeddings. For
instance, there is the issue of imbalanced classes -- many more negatives than positives -- when the link prediction in graphs is treated as a
classification problem~\cite{garciagasulla_2016}.  In the bioinformatics community the problem of imbalanced classes can be circumvented by
considering negative links that have a biological meaning, truncating thus many potential negative links that are highly improbable
biologically~\cite{alshahrani_2017}.  Other works have demonstrated that if no care is applied while splitting the datasets, we might end up producing
biased train and test examples, such that the implicit information from the test set may leak into the train set~\cite{toutanova_2016,dettmers_2017}. 
% In~\cite{dettmers_2017}, authors show that the random splits for the common knowledge graph evaluation benchmarks (Wordnet~\cite{miller_1995} and
% Freebase~\cite{bollacker_2008}) may bias the classification results for the symmetric relations.  Solutions to unbiased evaluations include curated
% data splits where no such information leakage is present. 
Kadlec et al.~\cite{kadlec_2017} have mentioned that the fair optimization of hyperparameters for competing approaches should be considered, as some
of the reported KG completion results are significantly lower than what they potentially could be. In the life sciences domain,
the time-sliced graphs as generators for train and test examples have been proposed as a more realistic evaluation benchmark~\cite{crichton_2018}, as
opposed to the randomly generated slices of graphs. 

In addition to reference implementations accompanying scientific papers that propose novel embedding methodologies, there is a wealthy core of open-source initiatives
that provide a one stop solution for efficient training of knowledge graph embeddings. For instance, pkg2vec~\cite{yu2019pykg2vec} and
PyKeen~\cite{ali_pykeen_2019} implement many of the state-of-the-art KG embedding techniques, with the focus on reproducibility and efficiency. While the community has
many options for the efficient KG embedding implementations, we believe that fewer attention has been drawn to evaluating neural embeddings when knowledge graphs may
exhibit structural changes. In this work we aim to make this gap narrower. Our work is closest in spirit to Goyal et al.~\cite{goyal2019} -- an evaluation
benchmark for graph embeddings that tries to explain which intrinsic properties of graphs make certain embedding models perform better. Unlike us, the authors consider
knowledge graphs with only one type of relation. 

The rest of this manuscript is organized as follows: in the Methods section (Section~\ref{sec:methods}) we define the notation, present knowledge graphs we used to
evaluate our approach, and formalize the evaluation pipeline. Then, in the Section~\ref{sec:structural-change} we introduce connectivity descriptors that allow us to
capture the structural change in knowledge graphs. Sections~\ref{sec:experiments},~\ref{sec:distmult-vs-complex} report our experiments and analysis. Finally, we conclude
our manuscript in the Section~\ref{sec:conclusion}.

\subsection{Contributions of this work}

In our work we define semantic similarity descriptors for knowledge graphs, and correlate the performance of neural embeddings to these descriptors. The big take away
from this work is that it is possible to improve the accuracy of the embeddings by adding more instances of semantically related relations. For instance, we can improve
overall accuracy of knowledge graph embeddings by increasing the number of semantically related relations. This means that if we have access to information that is
partially redundant (triples for an inferred relation, or a semantically related relation) this may improve overall accuracy. Moreover, by using our benchmark, we can
perform a more fine-tuned error analysis, by examining which specific type of links pose the most problem for overall link prediction. Finally, by examining the
correlation of accuracy scores to the semantic similarity descriptors we can explain the performance of neural embeddings, and predict their performance by simulating
modifications to the knowledge graphs.

% Our evaluation pipeline and the proposed semantic descriptors of knowledge graphs allows to make the following contributions to the community: i) show that there is no
% need to train graph embeddings separately for each relation in a knowledge graph, ii) 

%% file: sections/methods.tex
\section{Methods} \label{sec:methods}

\subsection{Notation and terminology}

Throughout this manuscript we employ the \emph{triple}-oriented treatment of knowledge graphs. As such, a knowledge graph $KG$ is simply a set of triples $(e_i, r_i, e_j)
\in KG$, where $e_i, e_j \in \ents(KG)$ are some entities, and $r_i \in \rels(KG)$ are its relation types, or simply relations. We assume that entities and relation types
are disjoint sets, i.e., $\ents(KG) \neq \rels(KG)$. Let $Pos_{KG}$ denote all the existing triples in the $KG$, i.e., triples $(e_i, r_i, e_j) \in KG$, and let
$Neg_{KG}$ denote the non-existing triples $(\bar{e_i}, \bar{r_i}, \bar{e_j}) \not \in KG$ via the knowledge graph completion ($\bar{e_i}, \bar{e_j} \in
\ents(KG)$, $\bar{r_i} \in \rels(KG)$). Similarly, $Pos_{r_i}$ and $Neg_{r_i}$ denote the existing and non-existing triples involving a relation $r_i$, respectively.
Obviously, in every triple of $Pos_{r_i}$ or $Neg_{r_i}$ the relation type is fixed to $r_i$. For each relation type $r_i$, $\dom(r_i), \range(r_i) \in \ents(KG)$
indicate the entities that belong to the domain and range of a relation $r_i$.  To describe the process of sampling some triples, we use the notation $\alpha X, \alpha
\in [0, 1]$, where $X$ is any set of triples. For instance, $(\alpha = 0.8) Pos_{r_i}$ is a sampled set of triples involving $r_i$, and consisting of $80 \%$ of triples
from $Pos_{r_i}$. Occasionally, when we write $(e_i, \_, e_j)$ we refer to a set of triples where $e_i, e_j$ are fixed, and the relation type $\_ \in \rels(KG)$ is free.

\subsection{Knowledge graphs} \label{sec:datasets}

We run our experiments on four different knowledge graphs: WN11~\cite{toutanova_2016} (subset of original WordNet dataset~\cite{miller_1995} brought down to 11 types of
relations, and without inverse relation assertions), FB15k-237~\cite{toutanova_2016} (a subset of Freebase knowledge graph~\cite{bollacker_2008} where inverse relations
have been removed), UMLS (subset of the Unified Medical Language System~\cite{bodenreider_2004} semantic network) and BIO-KG (comprehensive biological knowledge
graph~\cite{alshahrani_2017}). WN11, FB15k-237 and UMLS have been downloaded (December 2017) from the ConvE~\cite{dettmers_2017}
\footnote{\url{https://github.com/TimDettmers/ConvE}}{GitHub repository}, and BIO-KG has been downloaded (September 2017) from the official link indicated in the
\footnote{\url{http://aber-owl.net/aber-owl/bio2vec/bio-knowledge-graph.n3}}{supplementary material} for~\cite{alshahrani_2017}. Details on the derivation of subsets for
Wordnet and Freebase knowledge graphs can be found in~\cite{toutanova_2016,dettmers_2017}.

\subsection{Training and evaluating neural embeddings}

\begin{figure}[!h]
     \begin{center}
         \includegraphics[width=.99\textwidth]{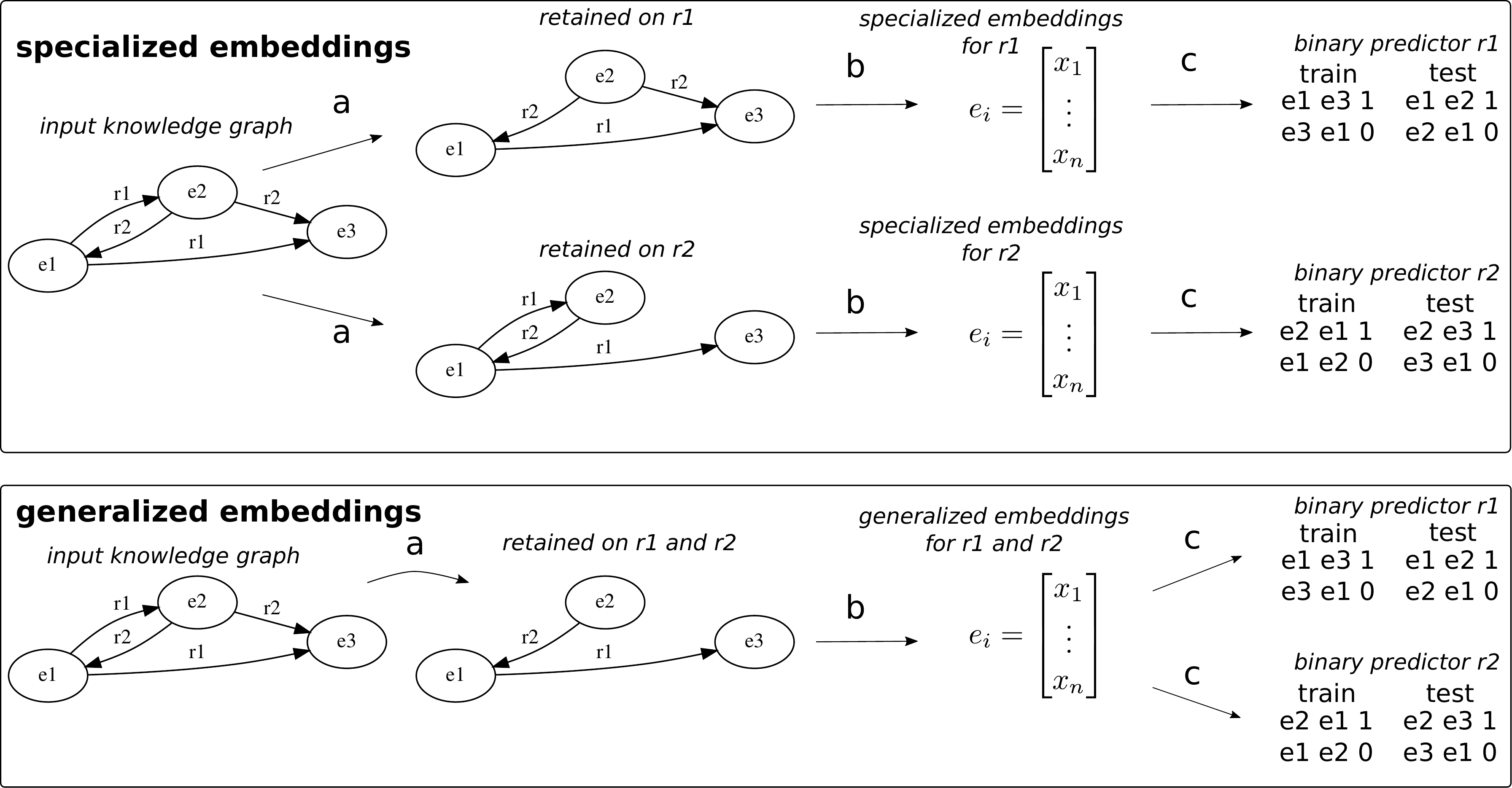}
     \end{center}
     \caption{\label{fig:pipeline-overview} Overview of the pipeline for training specialized and generalized neural embeddings for link prediction.}
\end{figure}

Our work builds upon the earlier proposed framework~\cite{alshahrani_2017} to both learn and evaluate neural embeddings for the knowledge graphs, which
we extend to make it more scalable. Throughout this manuscript we refer to the original framework as \emph{specialized embeddings} approach. In a nutshell this
approach learns and evaluates specialized embeddings for each relation type $r_i$ of entities of $KG$ as follows: a) we generate the retained graph
where we delete some of the triples involving $r_i$, then b) we compute the embeddings of the entities on this resulting retained graph, finally, c)
we assess the quality of these specialized embeddings on relation $r_i$ by training and testing binary predictors on positive and negative triples
involving $r_i$. These three steps are detailed in Figure~\ref{fig:pipeline-overview} in the \emph{specialized embeddings} box. The arrows labelled
with ``a'' in Figure~\ref{fig:pipeline-overview} symbolize the generation of retained graphs for relations $r_1$ and $r_2$, those marked with ``b''
computation of the entity embeddings, and ``c'' represents the training and testing binary classifiers for each relation type. 

The inconvenience of the specialized embeddings approach is that we need to compute entity embeddings for each relation type separately, which is a
serious scalability issue when the number of relation types in the knowledge graph becomes big. To circumvent this issue, we propose to train
\emph{generalized} neural embeddings for all relation types once, as opposed to training specialized embeddings for a specific relation $r_i$
(Figure~\ref{fig:pipeline-overview}, generalized embeddings box). Specifically, we generate only one retained graph, where we delete a fraction
of triples for each relation type $r_i$ (arrow marked with ``a'' on the bottom of Figure~\ref{fig:pipeline-overview}). This retained graph is then
used as a corpus for the computation of entity embeddings (``b''), which are then assessed with binary predictors for each relation type $r_i$ as in
the specialized case (arrows marked with ``c'' on the bottom of Figure~\ref{fig:pipeline-overview}). Evidently, this approach is more scalable and
economic, since we only compute and keep one set of entity embeddings per knowledge graph. 

In what follows we formalize the pipeline for link prediction with specialized and generalized neural embeddings, and we give a thorough description
of steps ``a'', ``b'' and ``c'' (Figure~\ref{fig:pipeline-overview}).

% To explain the correlations with the classifiers' performance, we introduce \emph{relation-centric} connectivity measures for the knowledge graphs to
% quantify these conditions. Additionally, in our analysis we take into consideration the ratios of missing examples during train and test phases. Our
% statistical validation tests the robustness of the neural embeddings, by evaluating their performances with different amounts of available information
% (i.e., percentage of available positive links), to simulate realistic scenarios where we have only limited data. We believe that the
% \emph{relation-centric} connectivity measures, the analysis of missed examples, and statistical validation with limited data are an important addition
% to the state-of-the-art evaluation toolbox for the knowledge graphs, which has not been proposed before. Our evaluation pipeline is formalized and
% made \footnote{\url{https://github.com/plumdeq/neuro-kglink}, Last accessed 2018-08-31}{open source}, and with this we aim to draw more attention of
% the community towards an important issue of transparency and reproducibility of the results.

\subsubsection{Generation of retained graphs (step a)}

\begin{figure}[!h]
    \begin{center}
        \includegraphics[width=.99\textwidth]{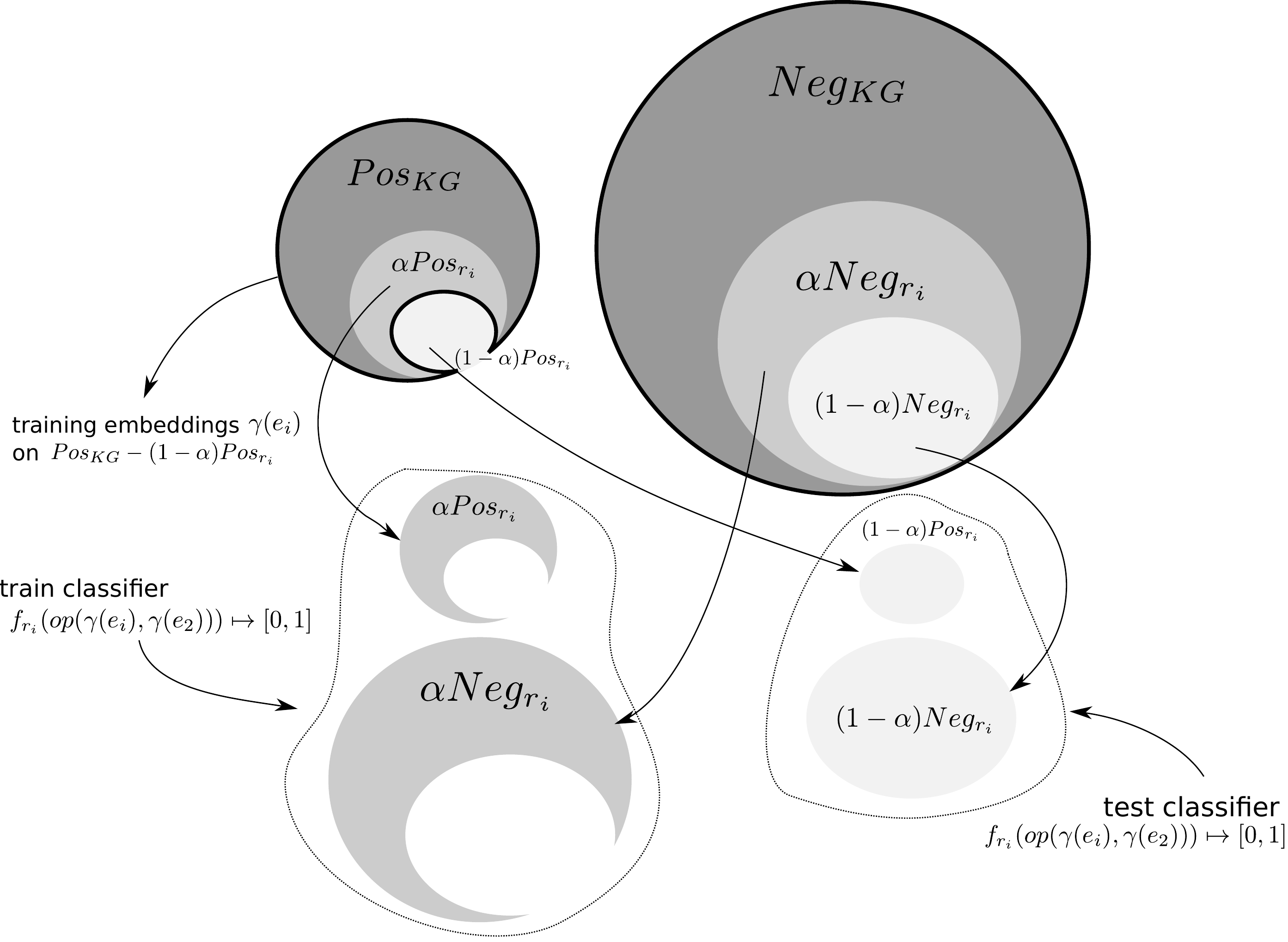}
    \end{center}
    \caption{\label{fig:evaluation} Schematic representation of the pipeline for the evaluation of the embeddings. $Neg_{KG}$ and its derivations (e.g., $\alpha
    Neg_{r_i}$) appear bigger visually to indicate that the elements are sampled from a much bigger set of all possible negative links.}
\end{figure}

By treating the problem of evaluation of the quality of the embeddings in a set-theoretic approach, we can define the following \emph{datasets}:

\begin{enumerate}
    \item $Pos_{KG} - (1-\alpha) Pos_{r_i}$ a specialized retained graph on $r_i$ -- training corpus for unsupervised learning of local to $r_i$ entity embeddings $\gamma_{r_i}(e_i)$ (in
        Figure~\ref{fig:evaluation} this set is demarcated with bold contour in the upper left corner),
    \item $Pos_{KG} - \bigcup_{r_i} (1-\alpha) Pos_{r_i}$ a generalized retained graph on all relations $r_i$ -- training corpus for unsupervised learning of global entity embeddings $\gamma(e_i)$,
    \item $\forall r_i, \alpha Pos_{r_i} \bigcup \alpha Neg_{r_i}$ -- train examples for the binary classifier for $r_i$,
    \item $\forall r_i,(1-\alpha) Pos_{r_i} \bigcup (1 - \alpha) Neg_{r_i}$ -- test examples for the binary classifier for $r_i$.
\end{enumerate}

\subsubsection{Neural embedding model (step b)} 

In this work we employ a shallow unsupervised neural embedding model~\cite{agibetov_semdeep_2018}, which aims at learning \emph{entity embeddings} in a dense $d$-dimensional
vector space. The model is simple and fast, and it embeds the entities that appear in the positive triples close to each other, and places the entities that appear in
negative triples farther appart. As in many neural embedding approaches, the weight matrix of the hidden layer of the neural network serves the role of the look-up matrix
(the matrix of embeddings - latent vectors). The neural network is trained by minimizing, for each positive triple $x_i = (e_i, \_, e_j)$ in the specialized ($x_i \in
Pos_{KG} - (1-\alpha) Pos_{r_i}$), or generalized graphs ($x_i \in Pos_{KG} - \bigcup_{r_i} (1-\alpha) Pos_{r_i}$), the following loss function

{\small
\begin{equation*}
    \sum_{(e_i, \_, \bar{e_j}) \in Neg_{KG}} L\left( \text{sim}(e_i, e_j) , \text{sim}(e_i, \bar{e_1}), \ldots, \text{sim}(e_i, \bar{e_k}) \right).
\end{equation*}
}

Where, for each positive triple $x_i$, we embed entities $e_i, e_j$ close to each other, such that $e_i$ stays as far as possible from the $k$ \emph{negative} entities
$\bar{e_1}, \ldots, \bar{e_k}$. The similarity function $\text{sim}$ is task-dependent and should operate on $d$-dimensional vector representations of the entities (e.g.,
standard Euclidean dot product). The loss function $L$ is a softmax, that compares the positive example ($e_i, e_j$) to all the negative examples 
(($e_i, x), x \in Neg_{KG}$).

\subsubsection{Link prediction evaluation with binary classifiers (step c)}

To quantify confidence in the trained embeddings, we perform the repeated random sub-sampling validation for each classifier $f_{r_i}$. That is, for each relation $r_i$
we generate $k$ times: retained graph $Pos_{KG} - (1-\alpha) Pos_{r_i}$ corpus for unsupervised learning of entity embeddings $\gamma_{r_i}(e_i)$) and train $\alpha
Pos_{r_i} \bigcup \alpha Neg_{r_i}$ and test $(1-\alpha) Pos_{r_i} \bigcup (1 - \alpha) Neg_{r_i}$ splits of positive and negative examples. Link prediction is then
treated as a binary classification task with a classifier $f_{r_i}: op(\gamma(e_i), \gamma(e_j))) \mapsto [0, 1]$, where $op$ is a binary operator
that combines entity embeddings $\gamma(e_i), \gamma(e_j)$ into one single representation of the link $(e_i, r_i, e_j)$. The performance of the classifier is measured
with the standard performance measurements (e.g., F-measure, ROC AUC). 

% We, in particular, report the F1 score (i.e., F-measure with equal weights for precision and recall).

%
% The pipeline for generation and evaluation of entity embeddings of a knowledge graph, presented in Section~\ref{sec:evaluation-embeddings}, has a big drawback that the
% embeddings for all entities in the knowledge graph ($\forall e_i \in KG, \gamma_{r_i}(e_i)$) are trained per relation $r_i$.  In the following, we refer to the approach
% where the embeddings are trained per each relation as \emph{local}, and we say \emph{global} when the embeddings are trained once, but evaluated separately for each
% relation $r_i$.  

% In this section we report on the possibility of training the embeddings $\forall e_i \in KG, \gamma(e_i)$ once on the retained graph $Pos_{KG} - \bigcup_{r_i} Pos_{r_i}$
% on all relations $r_i$, and evaluating it separately on all train $\alpha Pos_{r_i}$ and test examples $(1-\alpha) Pos_{r_i}$ for each relation $r_i$. 

\subsection{Evaluation benchmark summary and implementation}

The whole evaluation pipeline is summarized in Algorithm~\ref{alg:whole-pipeline}. In our experiments the specialized and generalized neural embeddings are trained with
the StarSpace toolkit~\cite{wu_2017} in the train mode 1 (see StarSpace specification) with fixed hyperparameters: embedding size $d=50$, number of epochs 10, all other
parameters set to default. Classification results are obtained with the scikit Python library~\cite{pedregosa_2011}, statistical analysis is performed with
Pandas~\cite{mckinney-proc-scipy-2010}. Our experiments were performed on a high performance cluster, with modern PCs controlling multiple NVIDIA GPUs (GTX1080 and
GTX1080TI). To demonstrate the high-flexibility of our pipeline, we also consider knowledge graph embeddings provided with the state-of-the-art
DistMult~\cite{distmult2015} and ComplEx~\cite{complex2016} models. Both of these models for our experiments were implemented in PyTorch (v1.2).

\begin{algorithm}[H]
    \caption{Evaluation of specialized and generalized knowledge graph embeddings}
	\label{alg:whole-pipeline}
	\begin{algorithmic}[1]
        \Require{$KG, \alpha, J$}
        \small
        \Statex{\footnotesize\emph{for each sub-sample validation run}}
        \For{$j \in 1, \ldots, J$} 
        \Statex{\footnotesize{\emph{\hspace{1em}generate retained graph on \textbf{all} $r_i$,}}}
        \Statex{\footnotesize{\emph{\hspace{1em}and compute \textbf{generalized} embeddings}}}
            \Let{$X$}{$Pos_{KG} - \bigcup_{r_i} (1-\alpha) Pos_{r_i}$} %{\footnotesize\Comment{retained graph $\forall r_i$}}
            \Let{$\gamma(e_i)$}{Embeddings($X$)} %{\footnotesize\Comment{generalized embeddings}}
            \Statex
            \For{$r_i \in \rels(KG)$}
                \Statex{\footnotesize{\emph{\hspace{2.5em}generate retained graph on $r_i$,}}}
                \Statex{\footnotesize{\emph{\hspace{2.5em}and compute \textbf{specialized} embeddings}}}
                \Let{$X_{r_i}$}{$Pos_{KG} - (1-\alpha) Pos_{r_i}$} %{\footnotesize\Comment{retained on $r_i$}}
                %\Statex{\footnotesize\Comment{compute specialized embeddings for $r_i$}}
                \Let{$\gamma_{r_i}(e_i)$}{Embeddings($X_{r_i}$)} 
                
                \Statex
                \Statex{\footnotesize\emph{\hspace{2.5em}generation of train/test examples for $r_i$}}
                \Let{$train_{r_i}$}{$\alpha Pos_{r_i} \bigcup \alpha Neg_{r_i}$}
                \Let{$test_{r_i}$}{$(1-\alpha) Pos_{r_i} \bigcup (1-\alpha) Neg_{r_i}$}
                %\Statex{\footnotesize\Comment{evaluate quality of specialized embeddings}}

                \Statex{\footnotesize\emph{\hspace{2.5em}evaluate quality of specialized embeddings}}
                \Let{$F1^j_{r_i}$}{$f_{r_i}(train_{r_i}, test_{r_i}, \gamma_{r_i}(e_i))$}
            \EndFor
            %\Statex
            %\Statex{\footnotesize\Comment{evaluate quality of generalized embeddings}}
            \Statex{\footnotesize\emph{\hspace{2.5em}evaluate quality of generalized embeddings}}
            \Let{$F1^j$}{$f_{r_i}(train_{r_i}, test_{r_i}, \gamma(e_i))$}
        \EndFor
        \Statex
        \Statex{\footnotesize\emph{average specialized embeddings evaluations}}
        \For{$r_i \in \rels(KG)$}
            \Let{$\widetilde{F1}_{r_i}$}{$\sum_j F1^j_{r_i}$}
        \EndFor
        \Statex
        \Statex{\footnotesize\emph{average generalized embeddings evaluations}}
        \Let{$\widetilde{F1}$}{$\sum_j F1^j$}
	\end{algorithmic}
\end{algorithm}

%% file: sections/structural_change.tex
\section{Structure of knowledge graphs and their change} \label{sec:structural-change}

In this section we introduce a few descriptors that are necessary to capture the variability and change in the structure of knowledge graphs. In addition to standard
descriptors that describe the structure of knowledge graphs syntactically (number of entities, relations and triples), we define descriptors that measure the positive to
negative ratio for each relation, and the semantic similarity of relations in the knowledge graph. These descriptors will be then used to evaluate and explain the
performance of neural embeddings. 

The variation of syntactic structure of the four graphs is summarized in Table~\ref{tab:datasets-stats} under the label \emph{Global}. By analyzing these global
descriptors, $|\ents|$, $|\rels|$, $|Pos_{KG}|$, we see that we have one small knowledge graph (UMLS), two medium-sized graphs (WN11, FB15K-237) and one very large
biological graph (BIO-KG). In what follows we define the descriptors that are used in the rest of the Table~\ref{tab:datasets-stats}.

\begin{table}[H]
\centering 
%\begin{adjustbox}{width=1\textwidth}
\scriptsize
    % \setlength\tabcolsep{10pt}
    \input{./tables/multi-relationness.tex}

%\end{adjustbox}
    \caption{\label{tab:datasets-stats}Statistics of descriptors that measure the variability in the structure of the four considered knowledge graphs. \emph{Global}
        descriptors describe knowledge graphs syntactically in terms of number of entities, relations and triples. Descriptors that measure the positive to
        negative ratio (\emph{Pos/Neg ratio}, see Equations~\ref{eq:mu}, \ref{eq:z}) are averaged for all relations, and show the percentage of positive to (semantic)
        negative examples. Finally, the semantic similarity of relations in the
    knowledge graphs (\emph{Semantic}) is summarized with the Frobenius norm on the Jaccard similarity matrices (Equations~\ref{eq:S}, \ref{eq:S'}).}
\end{table}

\subsection{Positive to negative ratio descriptors}

To measure the ratio of positive to negative examples in a knowledge graph, for a fixed relation $r_i$, we use the descriptors $\mu_{r_i}$ and $z_{r_i}$, defined as
follows:

{\small
\begin{align}
    \mu_{r_i} &= \frac{|Pos_{r_i}|}{|\dom(r_i)| \times |\range(r_i)|} \label{eq:mu}, \\
    z_{r_i} &= \frac{|Pos_{r_i}|}{|\ents| \times (|\ents|-1)}. \label{eq:z}
\end{align}
}

%Furthermore, we define 

\begin{figure}[!h]
    \begin{center}
        \includegraphics[width=.6\linewidth]{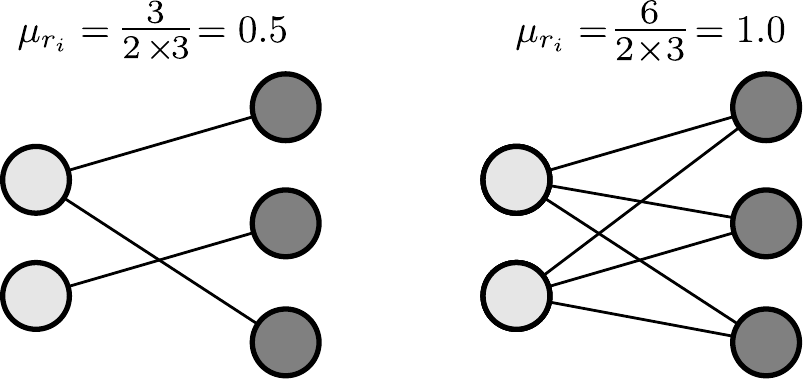}
    \end{center}
    \caption{\label{fig:measure-bipartedness}$\mu_{r_i}$ - ratio of positive to negative examples for the relation $r_i$. Effectively, it is maximized when the induced
        graph $G_{r_i}$ is closer to being complete (all entities are interconnected). Lighter nodes indicate \emph{domain}
    and darker nodes indicate \emph{range} of $r_i$.}
\end{figure}

For an induced graph $G_{r_i}$ -- consisting of all the triples involving relation $r_i$ of a knowledge graph -- both descriptors measure how close is $G_{r_i}$ to being
\emph{complete} (fully connected). Intuitively, if a graph is only half complete (Figure~\ref{fig:measure-bipartedness}, left), then we could potentially generate
as many negatives as positives. However, if the graph is complete (all entities are connected, Figure~\ref{fig:measure-bipartedness}, right), then there will be no
negative links generated. In $\mu_{r_i}$ we restrict the space of negative examples by generating semantically plausible links, i.e., we only consider unconnected pairs
from the domain and range of $r_i$. Analogously, $z_{r_i}$ relaxes this restrictions, i.e., the negatives can be generated from all possible pairs of entities in the
knowledge graph. We hypothesize that the performance of a binary link predictor of type $r_i$ should be positively correlated with both $\mu_{r_i}$ and $z_{r_i}$, i.e.,
the more training examples of type $r_i$ there are (the more connected $G_{r_i}$ is) the better is the performance of the binary predictor for $r_i$.

%connectedness of an induced graph $G_{r_i}$, 
%(Figure~\ref{fig:measure-bipartedness}). To this end, the maximum number of triples for relation $r_i$ is attained when all entities in the domain of the relation are
%connected to all the entities in the range (i.e., $\mu_{r_i} = 1$, see Figure~\ref{fig:measure-bipartedness}, right). Another interpretation of $\mu$ is the potential for
%the generation of negative examples, that is, low values of $\mu_{r_i}$ suggest that there might be many negative examples, i.e., the induced graph $G_{r_i}$ is highly
%incomplete. 

Focusing on the positive to negative ratio descriptors, under the label \emph{Pos/Neg ratio} in Table~\ref{tab:datasets-stats}, we see that we have two dense graphs (UMLS
and FB15k-237), and two very sparse graphs (WN11 and BIO-KG). When we restrict the negative sample generation to only plausible semantic examples ($\mu_{r_i}$), UMLS has
on average 60\% positive triples per relation, and FB15k-237 17.33\%. On the other hand, the two sparse graphs, BIO-KG, and WN11, both have less than 1\% positive
triples.  This suggests that for these sparse graphs, the binary prediction for any relation is extremely imbalanced, which may potentially hinder the performance of
neural embedding models. If we consider negative sample generation without any semantic restriction ($z_{r_i}$), then all binary tasks are highly imbalanced.

\subsection{Descriptors to measure semantic similarity of relations}

We introduce two descriptors that capture the amount to which the relations in the knowledge graph are similar one to another.  $S_{r_1, r_2}$ measures the number of
shared instances between the relations, and $S'_{r_1, r_2}$ measures the proportion of shared entities either in the domain or range of the two relations. Both are based
on the Jaccard similarity index, where sets are defined as in Equations~\ref{eq:S},~\ref{eq:S'}. Notice that $S_{r_1, r_2}$ can be seen as the degree of the role
equivalence in the description logic sense; the higher it is the more two relations are semantically similar (contain the same pairs of entities). And the $S'_{r_1, r_2}$
is higher when the two relations interconnect the same entities. Note that in $S$ elements of sets are tuples $(e_i, e_j)$, and in $S'$ elements are entities $e_i$.

{\small
\begin{align}
    S_{r_1, r_2} &= \frac{|Pos_{r_1} \bigcap Pos_{r_2}|}{|Pos_{r_1} \bigcup Pos_{r_2}|} \label{eq:S} \\
    S'_{r_1, r_2} &= \frac{|X(r_1) \bigcap X(r_2)|}{|X(r_1) \bigcup X(r_2)|}, X(r_i) = \dom(r_i) \cup \range(r_i) \label{eq:S'}
\end{align}
}

\begin{figure}[!h]
    \begin{center}
        \includegraphics[width=.99\textwidth]{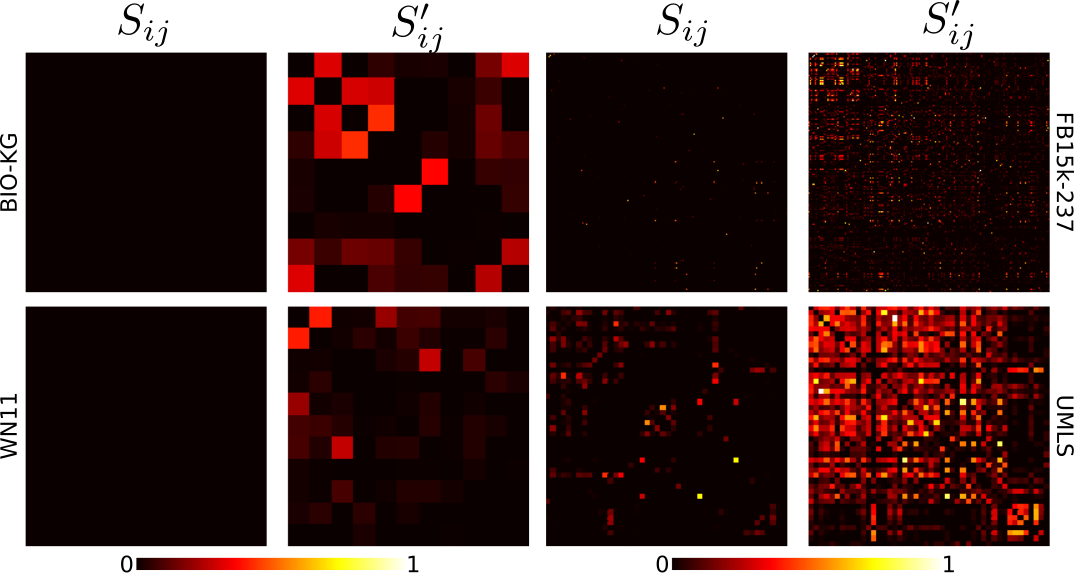}
    \end{center}
    \caption{\label{fig:relations-distributions}Heatmaps of semantic similarity among the relations in the four considered knowledge graphs. All matrices are symmetric
    and square, and have the size $|\rels| \times |\rels|$. $S_{ij}$ measures the semantic similarity as the Jaccard index on the shared instances (pairs of
entities), and $S'_{ij}$ measures the semantic similarity as the Jaccard index on the shared entities, in the domain or range of the two relations.}
\end{figure}

When we consider semantic similarity among the relations in our four knowledge graphs, we see the similar pattern as for the descriptors that measure positive to negative
ratio. In Figure~\ref{fig:relations-distributions} we demonstrate heatmaps of semantic similarity of relations for four graphs. 

We can observe that UMLS and FB15k-237 have many similar relations ($S_{r_1, r_2} \to 1$), very few of WN11 relations share instances, and the relations of BIO-KG do not
share any instances at all ($\forall_{r_1, r_2} S_{r_1, r_2} = 0$). If we consider the semantic similarity in terms of shared entities, we see that, although BIO-KG and
WN11 have dissimilar relations ($S_{r_1, r_2} \to 0$), they still can share information for the shared embeddings ($S'_{r_1, r_2} > 0$). To see it consider two relations
$r_1, r_2$ that do not share instances ($S_{r_1, r_2} \to 0$), but do share entities $e_1, \ldots, e_n$ that they interconnect ($S'_{r_1, r_2} > 0$). In this situation,
the training examples for $r_1$ and $r_2$ may share information during the learning phase and improve the quality of embeddings for $e_1, \ldots, e_n$. Using these two
similarity matrices, we can define measures of semantic similarity among relations in the whole knowledge graph by taking the Frobenius norm 
($\lVert S \rVert_F := \sqrt{\sum_{ij} |S_{ij}|^2}$) of matrices $S$ and $S'$ (Table~\ref{tab:datasets-stats}, label \emph{Semantic}).

Overall, the proposed connectivity descriptors capture well the semantic and structural variability of knowledge graphs, and will allow us to make more nuanced
evaluation of neural embeddings.

% \subsection{Impact of structural change on neural embedding learning phase}
% 
% We simulate the structural change by gradually adding links to the retained knowledge graphs. That is we use the $\alpha$ parameter to control the
% amount of available information during the generation of retained graphs. This allows us to benchmark the evolution of accuracy of neural embeddings,
% as we increase the available information ($\alpha \to 1$). However, in doing so one must also record the number of \emph{missed} embeddings due to the
% extreme deletion of links. To see it clearer, suppose that during the generation of the retained graph all triples $(e_k, \_,  \_) \in KG$ (only $e_k$
% fixed) involving entity $e_k$ are not assigned to the retained graph (i.,e.,  $(e_k,\_, \_) \not \in Pos_{KG} - (1-\alpha) Pos_{r_i}$), then the
% algorithm will not learn an embedding $\gamma(e_k)$, which will lead us to a situation where all the triples $(e_k, \_, \_) \in KG$ (examples at
% train/test) will be missing during training or testing of the binary classifier (depending whether these are assigned to to the train or test sets).
% 
% Intuitively, the amount of possible missing examples is inversely proportionate to the $\alpha$ parameter, i.e., the more information we include
% during the embedding learning phase, the fewer embeddings will be missed. We also hypothesize that the sparsely connected relations (i.e., sparsely
% connected graphs $G_{r_i}$ induced by the relation $r_i$) and the relations with very few examples (i.e., small $|Pos_{r_i}|$) will yield higher
% amounts of missed examples. 

%% file: tables/multi-relationness.tex
%\begin{tabular}{l|rrr|rrr|rr}
%\toprule
%                   & \multicolumn{3}{c}{Multi-relationness} & \multicolumn{3}{c}{Relation-based properties} & \multicolumn{2}{c}{Global properties} \\
%                   dataset & $\phi(KG)$                                          & $|Pos_{KG}|$                                    & $\frac{\phi(KG)}{|Pos_{KG}|}$ (\%) & $max_{|Pos_{r_i}|}$   & $min_{|Pos_{r_i}|}$ & $mean_{|Pos_{r_i}|}$ & $|\rels(KG)|$ & $|\ents(KG)|$ \\
%\midrule
%      WN11         & 124                                             & 93003                                           & 0.33 \%                                 & 37221  & 86    & 8459   & 11          & 40943 \\
% FB15k-237         & 23700                                           & 310116                                          & 7.42 \%                                 & 16391  & 45    & 1308   & 237         & 14541 \\
%      UMLS         & 1343                                            & 6527                                            & 20.76 \%                                & 1021   & 1     & 142    & 46          & 137 \\
%    BIO-KG         & 0                                               & 1619239                                         & 0.00 \%                                 & 554366 & 6159  & 179915 & 9           & 346225 \\
%\bottomrule
%\end{tabular}
%
\begin{tabular}{lrrrrrrr}
\toprule
            & \multicolumn{2}{c}{Semantic} & \multicolumn{2}{c}{Pos/Neg ratio} & \multicolumn{3}{c}{Global} \\
$G$         & $\lVert S \rVert_F$          & $\lVert S' \rVert_F$              & mean($\mu_{r_i}$) \%          & mean($z_{r_i}$) \% & $|\ents|$        & $|\rels|$    & $|Pos_{KG}|$   \\
\midrule
WN11        & 0.81                         & \textless 0.01                    & 0.68                          & 5e-4               & \emph{40,943}    & 11           & 93,003           \\
FB15k-237   & \textbf{19.09}               & \textbf{3.84}                     & \emph{17.33}                  & 6e-4               & 14,541           & \textbf{237} & \emph{310,116}           \\
BIO-KG      & 1.31                         & 0                                 & 0.73                          & 1e-4               & \textbf{346,225} & 9            & \textbf{1,619,239}           \\
UMLS        & \emph{9.46}                  & \emph{2.31}                       & \textbf{60}                   & \textbf{7e-1}      & 137              & \emph{46}    & 6,527           \\
\bottomrule
\end{tabular}

%% file: sections/experiments.tex
\section{Benchmarking specialized and generalized embeddings under structural change} \label{sec:experiments}

% \begin{algorithm}[!h]
%     \caption{Evaluation of specialized embeddings}
% 	\label{alg:specialized-embeddings}
% 	\begin{algorithmic}[1]
%         \For{$r_i \in \rels(KG)$}
%             \For{each sub-sample validation run} 
%                 \State{generate retained graph on $r_i$} \Comment{(a)}
%                 \State{generate train/test examples for $r_i$} \Comment{(a)}
%                 \State{compute embeddings $\gamma_{r_i}(e_i)$ for $r_i$} \Comment{(b)}
%                 \State{train and test binary predictor $f_{r_i}$} \Comment{(c)}
%             \EndFor
%             \State{average performance of $f_{r_i}$} \Comment{quality of $\gamma_{r_i}(e_i)$}
%         \EndFor
% 	\end{algorithmic}
% \end{algorithm}
% 
% 
% \begin{algorithm}[!h]
%     \caption{Evaluation of generalized embeddings}
% 	\label{alg:specialized-embeddings}
% 	\begin{algorithmic}[1]
%         \For{each sub-sample validation run} 
%             \State{generate retained graph on \textbf{all} $r_i$} \Comment{(a)}
%             \State{compute embeddings $\gamma(e_i)$ for \textbf{all} $r_i$} \Comment{(b)}
%             \For{$r_i \in \rels(KG)$}
%                 \State{generate train/test examples for $r_i$} \Comment{(a)}
%                 \State{train/test binary predictor $f_{r_i}$ on $\gamma(e_i)$} \Comment{(c)}
%             \EndFor
%         \EndFor
%         \State{average performance of $f_{r_i}$} \Comment{average quality of $\gamma(e_i)$ for each $r_i$}
% 	\end{algorithmic}
% \end{algorithm}

%
The goal of our experiments is to empirically investigate if, and when, the generalized neural embeddings attain similar performance as the specialized embeddings, for
the four considered datasets. To do so, we first generate the retained graphs, and the train and test datasets. The retained graphs are generated for each relation type
$r_i$ in the case of specialized embeddings, and only once for the generalized embeddings. We always keep the 1:1 ratio for the positive and negative examples. When we
sample the negatives for a relation $r_i$, we only consider the triples $(\bar{e_i}, r_i, \bar{e_j})$ where the entities come from the domain ($\bar{e_i} \in \dom(r_i)$)
and the range ($\bar{e_i} \in \range(r_i)$) of $r_i$. Embeddings are computed from retained graphs, and then evaluated on train and test datasets. Note that
we only provide results for generalized embeddings for FB15k-237, since the computation of specialized embeddings for 237 relations of FB15k-237 would take months
(on our machine) to finish\footnote{computation of specialized embeddings grows linearly with the number of relations, and exponentially in the number of repeated sub-sample
validation runs}{}. The evaluation of embeddings for one relation type $r_i$ is performed with the logistic regression classifier $f_{r_i}(op(\gamma (e_i), \gamma
(e_j)))$, where $op$ is the vector concatenation operator. To test the robustness of embeddings we perform evaluations with limited information, i.e., the size of
the retained graphs controlled by $\alpha \in \{0.2, 0.5, 0.8\}$, and we analyze the amount of missed embeddings in all experiments. All of our results are presented as
averages of 10 repeated random sub-sampling validations. We thus report mean F-measure scores and their standard deviations. 

\subsection{Comparing accuracy}

\begin{figure}[!h]
    \begin{center}
        \includegraphics[width=.99\textwidth]{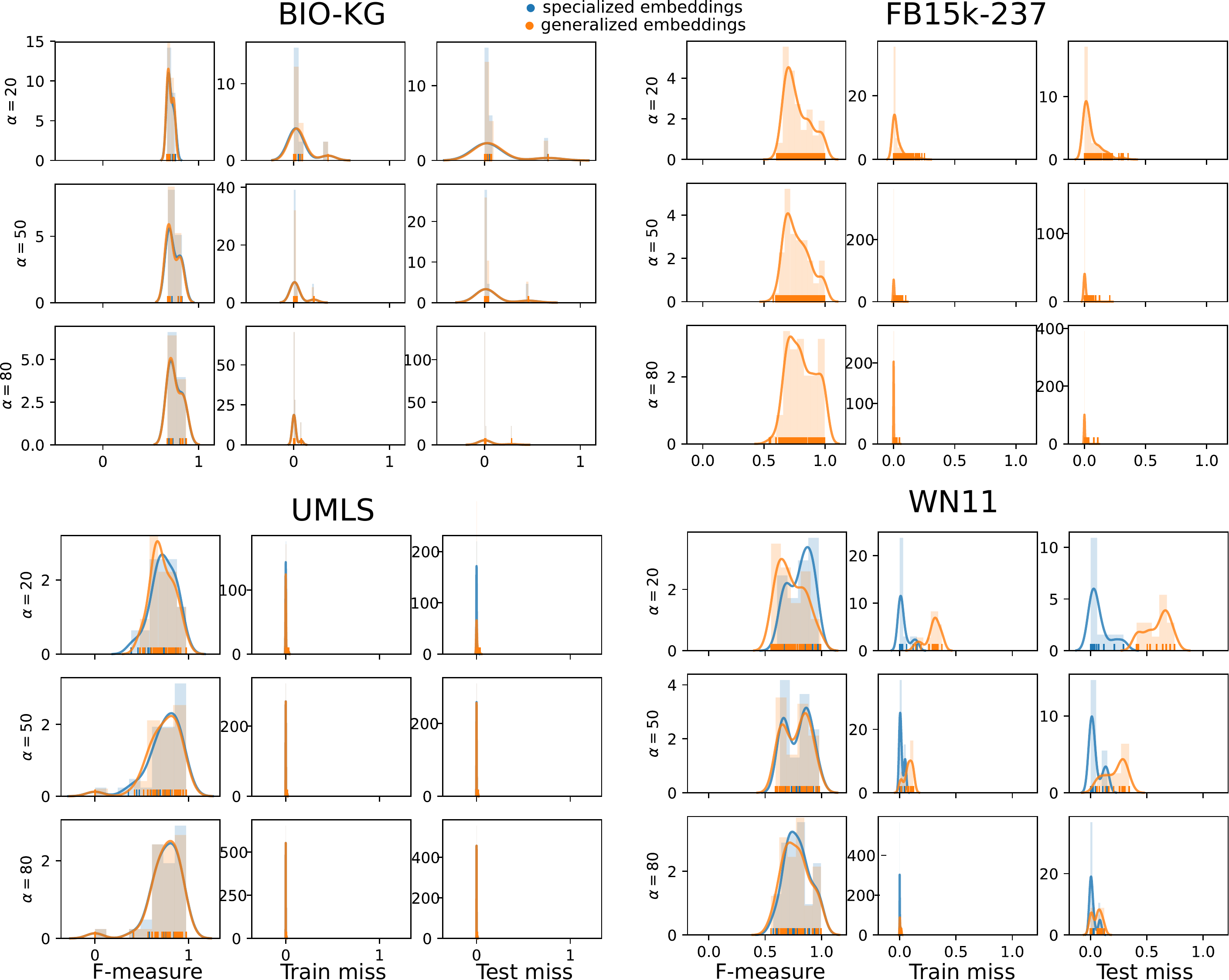}
    \end{center}
    \caption{\label{fig:f-mean-miss-distributions}Distributions of F1 scores, ratio of missing entities at trainging, and at testing, for specialized and generalized embeddings, for four graphs.}
\end{figure}

In Figure~\ref{fig:f-mean-miss-distributions} we present distributions of averaged F1 scores, which measure the accuracy of embeddings, and ratios ($[0,
1])$ of missed examples at training and testing of the binary classifiers $f_{r_i}$. As such, the overall performance of specialized or generalized embeddings on
one knowledge graph is characterized by these three distributions over all relations in the given knowledge graph. The performance of embeddings is compared with varying
amount of information present at the time of training of neural embeddings (parameter $\alpha$). All distributions in Figure~\ref{fig:f-mean-miss-distributions} are
estimated and normalized with kernel density interpolation from actual histograms.  

In the three knowledge graphs: BIO-KG, UMLS and WN11, distributions converge as we increase the amount of available information (e.g., $\alpha \to 1$), which supports
of the hypothesis of this manuscript, that the generalized embeddings may yield the similar (if not the same) performance as specialized embeddings. When we
consider BIO-KG, the F1 and missing examples distributions for specialized and generalized neural embeddings converge almost to identical distributions, even when the
overall amount of information is low (e.g., only 20 \% of available triples). This may be explained by a relatively big size of available positive examples per relation
type $|Pos_{r_i}|$ (hundreds of thousands of available triples per relation). Though, $\text{mean}(z_{r_i})$ and $\text{mean}(\mu_{r_i})$
(Table~\ref{tab:datasets-stats}) are very similar for BIO-KG and WN11, differences between specialized and generalized embeddings for WN11 are much more
characterized, than in the case of BIO-KG. In particular, neural embeddings for WN11 are very sensitive to $\alpha$, the less information there is the more is the
intra-discrepancy in specialized and generalized distributions for the same scores (F1 and the ratio of missed examples). The amount of missed examples is very high
for both specialized and generalized cases, for smaller values of $\alpha < 0.8$, and distributions converge when $\alpha=0.8$. The most regular behavior is
demonstrated by neural embeddings trained on UMLS corpora, where missing examples rates are all almost zero, even when $\alpha=0.2$. Shapes of the F1
distributions are very similar for all values of $\alpha$, intra-discrepancies are very low. These observations allow us to hypothesize that similar trends
might exist for the FB15k-237 knowledge graph, since UMLS and FB15k-237 have similar distributions of $\mu_{r_i}$ and $z_{r_i}$. 

To summarize, as we increase the amount of available information during the computation of neural embeddings ($\alpha \to 1$) intra-discrepancies
between specialized and the generalized embeddings become negligible. And this is good news, since training generalized embeddings is
$|\rels(KG)|$-times faster than training specialized embeddings for each relation $r_i$, with the strong evidence that if we have enough
information we can achieve the same performance.

\subsection{Comparing average performance}

\begin{figure}[!h]
    \begin{center}
        \includegraphics[width=.6\linewidth]{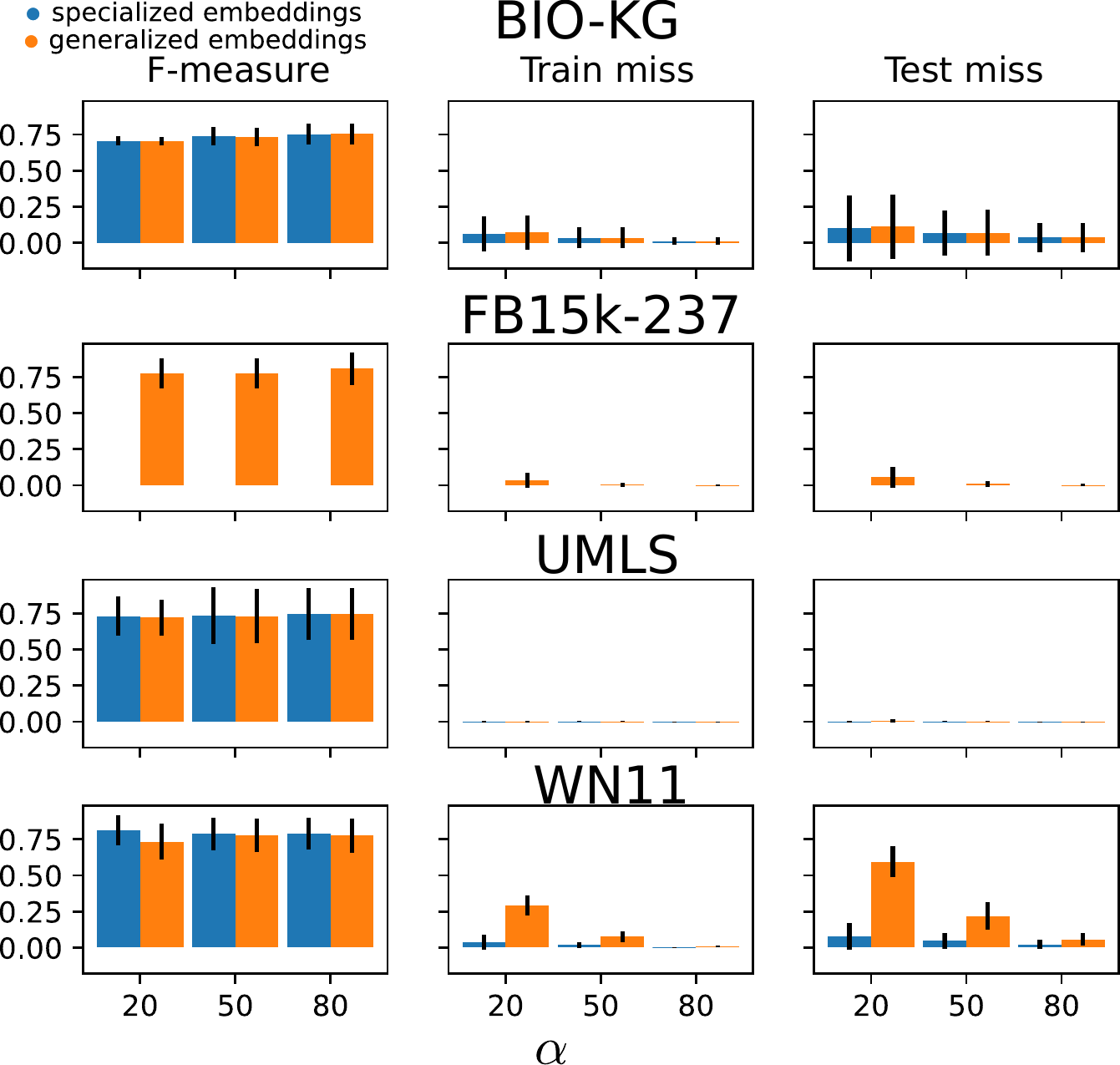}
    \end{center}
    \caption{\label{fig:averaged-metrics}Averaged F1 scores, ratio of missing entities at trainging, and at testing, for specialized and generalized embeddings, for four graphs.}
\end{figure}

We recall that each distribution's sampled point is obtained by averaging results of $k$ repeated experiments for one relation $r_i$. To directly
compare distributions, we compare their means and standard deviations, and, as such, we are comparing the average performance of $|\rels(KG)|$ binary
classifiers for specialized and generalized neural embeddings, with the varying parameter $\alpha$. Figure~\ref{fig:averaged-metrics} depicts the
average performance of all binary classifiers and its standard deviation for the four knowledge graphs. 

As expected, the performance of specialized embeddings is better than the performance of generalized embeddings, however differences are very slim. BIO-KG and UMLS
demonstrate that, as we increase $\alpha$, the average F1 score increases in both cases, however, so does the standard deviation as well. WN11, on the other hand,
demonstrates a counter-intuitive trend where the best performance of specialized embeddings occurs when less information is available. And, for specialized embeddings,
the F1 score decreases slightly when we include more information during the computation of neural embeddings. This maybe explained by an increased amount of missing
examples, both during training and testing of the binary classifier. Due to a very sparse connectivity of the induced graphs $G_{r_i}$ of WN11, when we only consider $20
\%$ of available triples -- we exclude 80 \% of available links -- many entities are likely to become disconnected. This means that no embeddings are learned for them,
and, as a result, the binary classifier is both trained and tested on fewer examples.

\section{DistMult vs. ComplEx} \label{sec:distmult-vs-complex}

\begin{figure}[!h]
    \begin{center}
        \includegraphics[width=.6\linewidth]{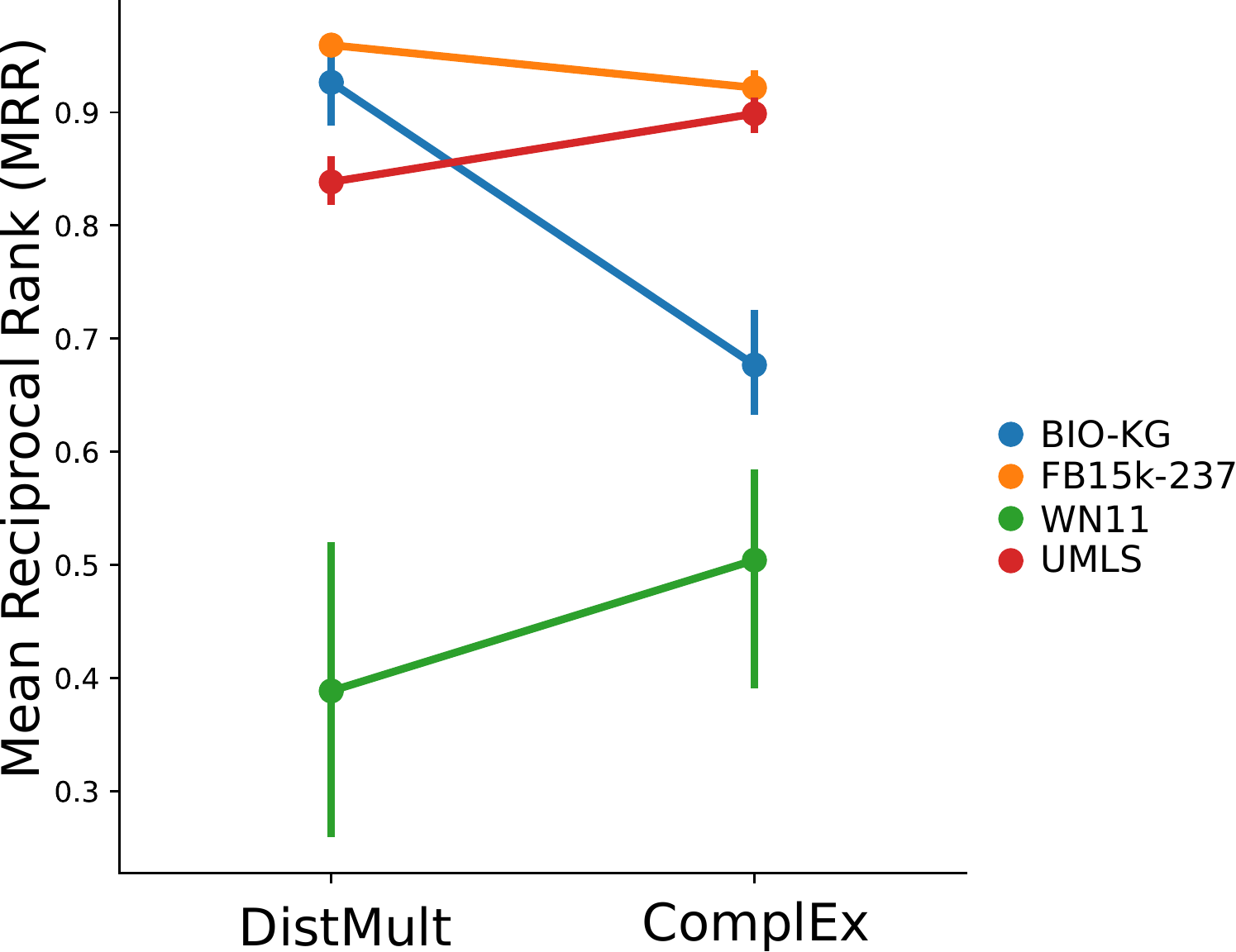}
    \end{center}
    \caption{\label{fig:point-plot-distmult-vs-complex}Point estimate of performances for the two models with confidence intervals. Horizontal lines accentuate the
    differences for the same knowledge graph.}
\end{figure}

In this experiment, our goal is to compare two of the most popular knowledge graph embedding models, DistMult~\cite{distmult2015} and ComplEx~\cite{complex2016}, by using
our relation-centric ablation benchmark. In particular, our mission is to explain which intrinsic properties of graphs directly impact the accuracy of neural models. In
contrast to our previous experiment, we perform random ablations on each relation type $\alpha_r \sim N(0, 1), \forall r \in \rels(KG)$. For each knowledge graph we train
10 DistMult and 10 ComplEx models. We fix the embedding dimension to 200, and we use Adam optimizer. Each time models are trained for 50 epochs. The accuracy is assessed
with mean rank and mean reciprocal rank metrics. In Table~\ref{tab:distmult-vs-complex-global} we report mean (with standard deviation) MR and MRR performance of two
models on four datasets, as well as averaged performance of models for all graphs. Overall, ComplEx is slightly better than DistMult ($0.76 (0.2)$ \emph{vs.} $0.75
(0.26)$, mean(SD)) however their performances stay within the confidence intervals for all graphs. If we look at the performance of models for specific graphs, the
differences are more apparent. In Figure~\ref{fig:point-plot-distmult-vs-complex} we present point estimates and confidence intervals of the MRR metric for a specific
graph, with horizontal lines accentuating the difference in accuracy for the two models. ComplEx is better at UMLS and WN11. DistMult on the other two. 

\begin{table}[H]
\centering 
%\begin{adjustbox}{width=1\textwidth}
\small
    \input{./tables/dm-complex-per-model.tex}
%\end{adjustbox}
    \caption{\label{tab:distmult-vs-complex-global}Average performance of DistMult and ComplEx on four knowledge graphs. MRR and MR are reported as means of 10 runs
    (standard deviation in parenthesis). Additionally, we report average performance per model for all knowledge graphs (with standard deviation).}
\end{table}

\begin{figure}[!h]
    \begin{center}
        \includegraphics[width=.8\linewidth]{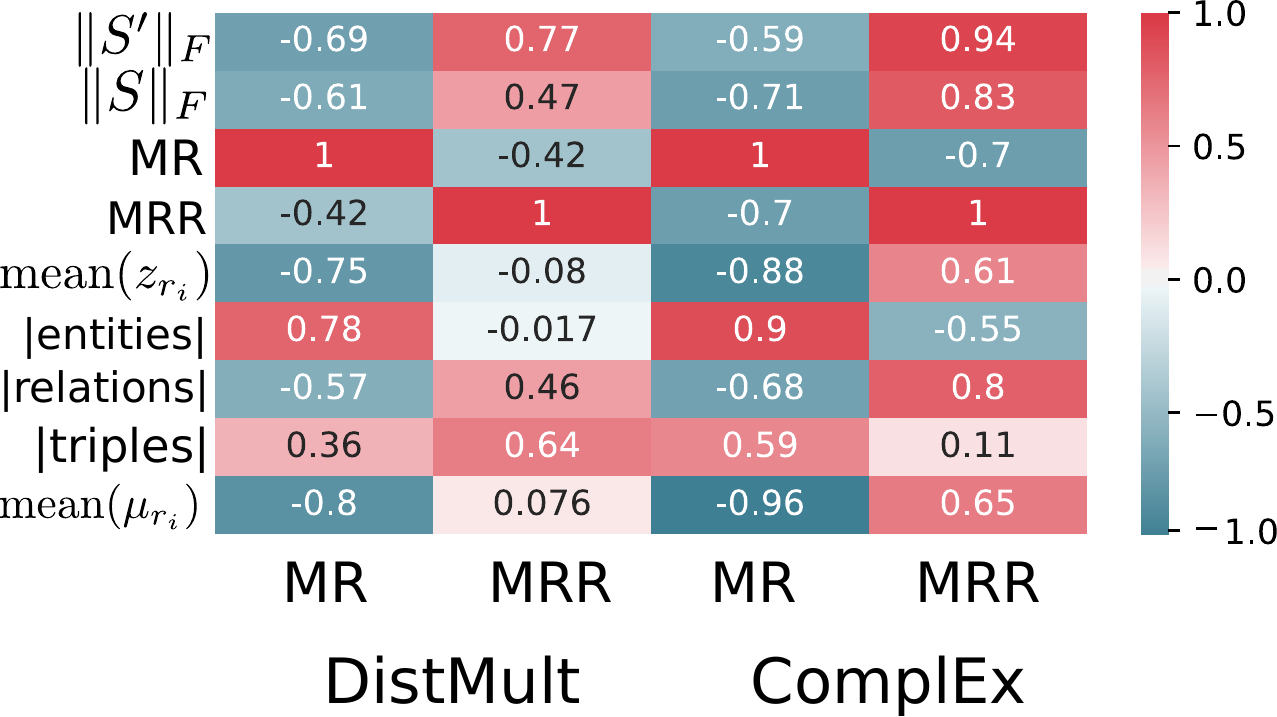}
    \end{center}
    \caption{\label{fig:spearman-distmult-vs-complex}Spearman correlation of accuracy performance to structural descriptors of knowledge graphs.}
\end{figure}

\begin{figure}[!h]
    \begin{center}
        \includegraphics[width=1.\linewidth]{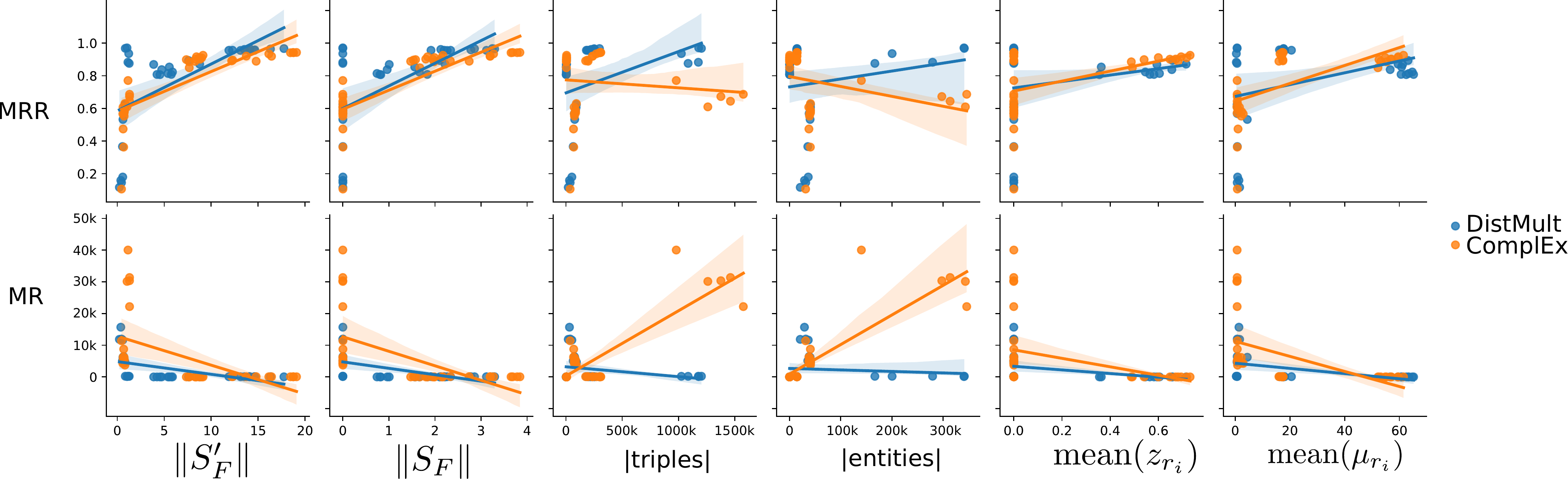}
    \end{center}
    \caption{\label{fig:regplot-distmult-vs-complex}Regression plots detailing the linear correlation of the performance of a model to structural descriptors of knowledge
    graphs.}
\end{figure}

To explain such a disparity in performance we analyze correlations of model's performance to intrinsic properties of (test) graphs.
Figure~\ref{fig:spearman-distmult-vs-complex} summarizes Spearman correlations, and the Figure~\ref{fig:regplot-distmult-vs-complex} shows regression plots to emphasize
correlations.

In the following we use the abbreviation \emph{corr} to refer to the Spearman correlation (since distributions are potentially not normal) of the MRR metric to intrinsic
properties of graphs (Figure~\ref{fig:spearman-distmult-vs-complex}). We see that ComplEx depends on properties that emphasize semantic similarity among the relations
(corr to $\lVert S' \rVert_F \text{ at } 0.94$ and to $\lVert S \rVert_F \text{ at } 0.83$), it performs better whenever the graph has semantically related relations (be it dense or sparse), and it depends
less on the number of triples (samples, corr 0.11), suggesting that this model better learns semantic relationships within the graph. On the other hand, DistMult
leverages less the role equivalence similarity (corr to $\lVert S \rVert_F = 0.47$), concentrates more on similar entities (corr to $\lVert S' \rVert_F = 0.77$), and is highly sensitive to the
number of triples (corr to $\sum_{r_i} |Pos|_{r_i}$ 0.64). This may explain why ComplEx outperforms DistMult at a small and extremely dense UMLS graph, and at a
relatively big and sparse WN11 graph. DistMult, on the other hand, better leverages the abundant presence of triples in the big and sparse BIO-KG graph, and in the
dense FB15k-237 graph. By looking at the regression plots (Figure~\ref{fig:regplot-distmult-vs-complex}), we can see that both models have high variability (small
confidence) for the graphs that exhibit low semantic similarity among the relations, and contain very few samples at their disposal. Overall, the ComplEx model is better
at extracting semantic relationships, while DistMult is better at leveraging big sample sizes. 

%% file: tables/dm-complex-per-model.tex
\begin{tabular}{llrrrr}
\toprule
                  &           & MRR (mean (SD))      & MR (mean (SD)) \\
model             & $G$       &                      & \\
\midrule
ComplEx           & BIO-KG    & 0.67 (0.06)          & 30,810.75 (6330.92) \\
                  & FB15k-237 & 0.92 (0.02)          & \textbf{63.56 (20.61)} \\
                  & WN11      & \textbf{0.5 (0.15)}  & \textbf{6,033.56 (2360.53)} \\
                  & UMLS      & \textbf{0.89 (0.02)} & \textbf{2.87 (0.36)} \\
\cmidrule{3-4}
                  &           & \textbf{0.76 (0.2)}  & 6144.39 (10,846.1) \\
\cmidrule{2-4}
DistMult          & BIO-KG    & \textbf{0.92 (0.04)} & \textbf{189.26 (42.09)} \\
                  & FB15k-237 & \textbf{0.95 (0.01)} & 67.96 (14.48) \\
                  & WN11      & 0.38 (0.21)          & 8,347.39 (4016.02) \\
                  & UMLS      & 0.83 (0.03)          & 4.27 (0.78) \\
\cmidrule{3-4}
                  &           & 0.75 (0.26)          & \textbf{2,432.64 (4321.82)} \\
\bottomrule
\end{tabular}

%% file: sections/conclusion.tex
\section{Conclusions} \label{sec:conclusion}

The lessons learned from our experiments lead us to conclude that neural embeddings' performance depends on the degree of how tight the relations within the knowledge
graph interconnect entities. The presence of multiple relations -- edges -- that make the overall spider web of entities more entangled, affect the accuracy.
Therefore, to increase the accuracy of neural embeddings in knowledge bases we would identify two main ingredients: a) increase the sample size, b) add similar relations.
Obviously, by introducing novel relation types we increase the overall sample size. The addition of semantically similar relations can be achieved by using logical
reasoners, or by recurring to external data sources. For instance, language models could be used to augment knowledge bases~\cite{petroni2019}.

Herein, we proposed an open-source evaluation benchmark for knowledge graph embeddings that better captures structural variability and its change in real world knowledge
graphs. 

%Connectivity descriptors introduced in this work allow for a more thorough and nuanced benchmarking of embeddings for their accuracy and
%robustness. We applied this benchmark to compare specialized and generalized neural embeddings and found that the performance distributions converge
%even with very limited available information ($20 \%$ of all positive triples), provided that the knowledge graph is dense and exhibits uniform
%connectivity over all relation types. However, the performance distributions would need more available information to converge, for the knowledge
%graphs with sparse connectivity and the low number of positive triples per relation type. Such an evaluation benchmark will be especially useful in
%the situations where we want to simulate the evolution of accuracy of embeddings for the knowledge graphs that are expected to grow in size. Finally,
%we make our code open source, and hope that it will contribute to the standardization of knowledge graph embedding evaluation benchmarks.